\documentclass[11pt,a4paper]{article}
\usepackage[hyperref]{acl2020}
\usepackage{times}
\usepackage{latexsym}
\usepackage{multicol}
\usepackage{multirow}
\usepackage{algorithm}
\usepackage[noend]{algpseudocode}
\usepackage[colorinlistoftodos]{todonotes}
\usepackage{comment}
\usepackage{enumitem}
\usepackage{breqn}
\usepackage{tabularx}
\usepackage{arydshln}
\usepackage{graphicx}

\setlist[itemize]{noitemsep, topsep=0pt}

\usepackage{array}
\newcolumntype{L}[1]{>{\raggedright\let\newline\\\arraybackslash\hspace{0pt}}m{#1}}
\newcolumntype{C}[1]{>{\centering\let\newline\\\arraybackslash\hspace{0pt}}m{#1}}
\newcolumntype{R}[1]{>{\raggedleft\let\newline\\\arraybackslash\hspace{0pt}}m{#1}}

\usepackage{url}

\aclfinalcopy % Uncomment this line for the final submission
\def\aclpaperid{993} %  Enter the acl Paper ID here

\title{Simple-QE: Better Automatic Quality Estimation \\
for Text Simplification}

\author{Reno Kriz \\
  University of Pennsylvania \\
  \texttt{rekriz@seas.upenn.edu} \\\And
  Marianna Apidianaki \\
  University of Helsinki, Finland \\
  \texttt{marianna.apidianaki@helsinki.fi} \\\And
  Chris Callison-Burch \\
  University of Pennsylvania \\
  \texttt{ccb@cis.upenn.edu} \\}

\author{\parbox{12cm}{\centering Reno Kriz$^{*}$, Marianna Apidianaki$^{\triangle}$, and Chris Callison-Burch$^{*}$} \\
$^{*}$ Computer and Information Science Department, University of Pennsylvania \\
$^{\triangle}$ CNRS, LLF, France and University of Helsinki, Finland \\
{\tt \{rekriz,ccb\}@seas.upenn.edu}, {\tt marianna.apidianaki@helsinki.fi}
}

\date{}

\begin{document}
\maketitle

\begin{abstract}
    Text simplification systems generate versions of texts that are easier to understand for a broader audience. 
    The quality of simplified texts is generally estimated using metrics that compare to human references, which can be difficult to obtain. 
    We propose Simple-QE, a BERT-based quality estimation (QE) model adapted from prior summarization QE work, and show that it correlates well with human quality judgments.  
    Simple-QE does not require human references, which makes the model useful in a practical setting where users would need to be informed about the quality of generated simplifications.
    We also show that we can adapt this approach to accurately predict the complexity of human-written texts.
    %\textcolor{red}{make sure to add sth about the IR experiment, if you keep it}
    %correlates well with human judgments on the fluency, adequacy, and complexity of simplification system outputs. We also show that fine-tuning BERT to capture the complexity of human-written text works extremely well. 
    %attempt to make texts easier to understand for a broader audience.
    %Automatic metrics such as SARI have generally been used when evaluating sentence simplification systems.
    %However, these metrics require one or more reference simplifications, which can be both expensive to obtain and limiting when there are many reasonable simplifications.
    %We present Simple-QE, a BERT-based quality estimation model that correlates well with human judgments on the fluency, adequacy, and complexity of simplification system outputs.
\end{abstract}

\section{Introduction}

Simplification systems make texts easier to understand for people with reading disabilities and language learners, or for readers not yet familiar with a specific domain. They propose re-writings using simpler, meaning-preserving words and structures. For automatically-produced simplifications to be useful in practical settings, users should be able to easily assess their quality. 
%and whether they can be trusted. %We propose a model for measuring the quality of automatically generated simplifications without need for human references. 
%The quality of automatic simplifications varies, and ways for assessing whether the generated text can be trusted are needed. 
Traditional evaluation metrics \cite{papineni2002bleu,xu2016optimizing} estimate the quality of generated texts by comparing them to human-written simplifications, which restricts their use to settings where such references are available. 
%When a user wishes to access   automatically simplified text, the quality of the latter cannot be estimated using such metrics. 
In addition, comparing simplifications to a single reference is often too restrictive, as most texts can be simplified in a variety of ways.

We propose a model for measuring the quality of automatically generated simplifications, which does not require human references. Our model, Simple-Quality Estimation (Simple-QE), adapts the BERT-based summary QE model Sum-QE \cite{xenouleas2019sumqe}, to the simplification setting. 
% and enriches it with features that are relevant for estimating the quality of simplifications. 
As opposed to summaries -- which contain specific pieces of information from the original text(s) %judged as important, 
and omit unimportant passages -- simplified text typically expresses all the content present in the original text using simpler words and structures. 
%that can be more easily understood. 
Both types of text, however, need to fulfill some linguistic quality constraints in order to be useful, such as being  grammatical and well-formed. We show that Simple-QE correlates well with human judgments of linguistic quality on system output %outputs 
produced by simplification systems. In addition, we %can 
adapt our model to make reasonable complexity predictions at both the sentence and document level. Our models can be used to optimize both simplification system development and the process of writing manual simplifications.

\begin{figure*}[bt]
    \centering
    \includegraphics[width=11.5cm]{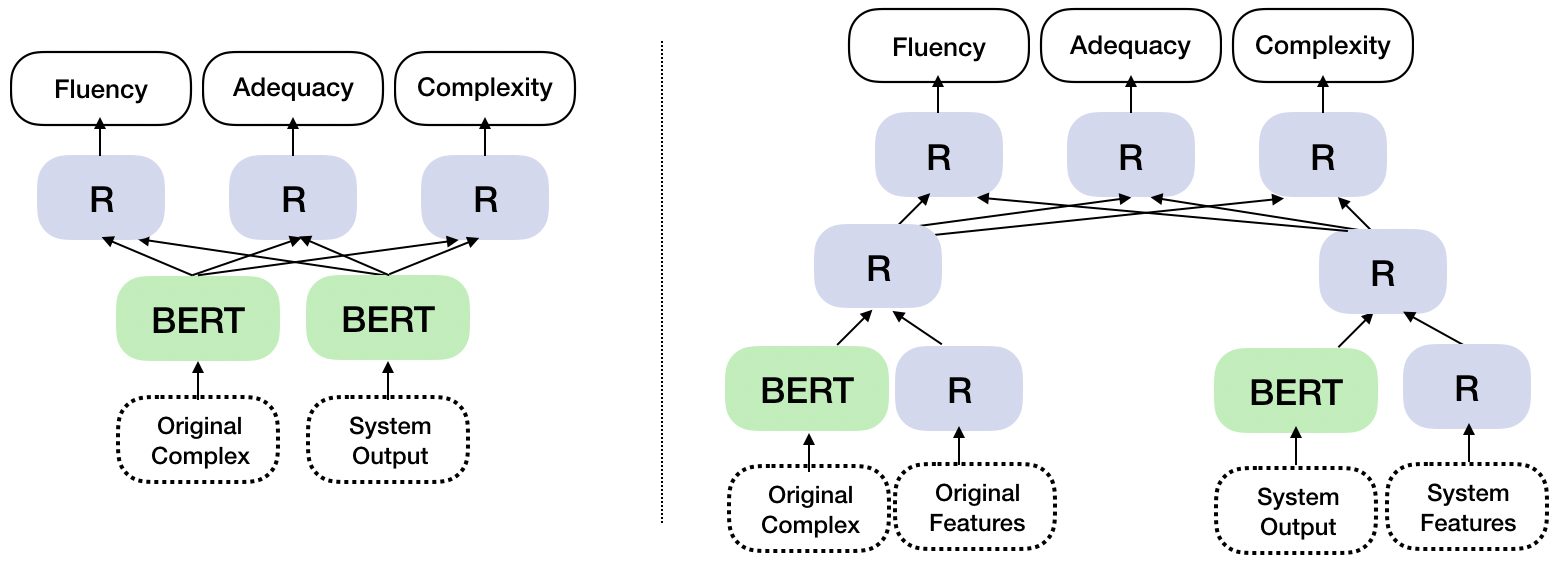}
\caption{The Simple-QE architecture, with and without adding numeric features corresponding to the original complex sentence (Original Features) and the system output (System Features). $R$ denotes a regressor layer.}
\label{architecture}
\end{figure*}

\section{Related Work}

%\subsection{Quality Estimation Methods}

Quality Estimation (QE) methods were first introduced in the field of machine translation %(MT) 
to measure the quality of automatically translated text without need for reference translations  \cite{bojar2017findings,martins2017pushing,specia2018findings}. In the most recent QE task, the best systems leveraged BERT via pre-training for specific language pairs and integrating a transfer learning approach \cite{fonseca2019findings}. %The goal of these models is to measure the quality of text automatically translated in different languages without the need for reference translations, as these can be expensive to obtain. %The most comparable work to ours,

\newcite{xenouleas2019sumqe} propose several extensions to the BERT fine-tuning process \cite{devlin2019bert} to estimate summary quality. %summarization output. 
Their proposed model, Sum-QE, %which fine-tunes BERT to 
predicts five linguistic qualities of generated summaries using multi-task training: 
Grammaticality, Non-redundancy, Referential Clarity, Focus, and Structure and Coherence. Similar to  state-of-the-art results obtained by BERT on many classification tasks, %(including entailment prediction, sentiment analysis, and textual similarity), 
\newcite{xenouleas2019sumqe} show that BERT can be successfully applied to QE.
%Grammaticality, Non-redundancy, Referential Clarity, Focus, and Structure \& Coherence. 
%Our goal is to make it possible to estimate the quality of simplified texts without need for model references, which is the main advantage of QE models compared to standard evaluation metrics.
In our work, we adapt Sum-QE to simplification QE, estimating the Fluency, Adequacy and Complexity of simplified text. %\textcolor{gray}{Sum-QE can estimate the well-formedness of summaries and simplifications by only looking at the system output. To assess adequacy and complexity, in our experiments we also include the original complex sentence along with the simplified system output during training and testing. This results in significant increases in correlations with  human judgments of adequacy and relative complexity.}
%The Sum-QE model is based on BERT, and .. ++ In our work, we adopt a transfer learning approach and adapt the Sum-QE model to simplification. 

Most recent simplification research %evaluate the quality of the proposed models using 
uses %a combination of 
both automatic metrics and human judgments during evaluation  \cite{zhang2017sentence,kriz2019complexity,mallinson2019controllable}. The %automatic 
metrics commonly used %for optimizing and evaluating sentence simplification systems
are BLEU \cite{papineni2002bleu}, METEOR \cite{banerjee2005meteor}, and SARI \cite{xu2016optimizing}. Contrary to Simple-QE, these metrics require a reference sentence. 
%restricting their application to specific usage settings. 
%\textcolor{gray}{Note that these are not directly comparable to this work, because they require a reference sentence.}
%BLEU has also been widely used for evaluating generation output, 
%Moreover, recent work has shown that 
%Beyond this, 
Furthermore, BLEU correlates poorly with deletion, a core aspect of simplification \cite{sulem2018bleu}. SARI correlates well with human simplification judgments at the lexical level, 
%human judgments about simplifications at the lexical level, 
but this does not transfer to the sentence level, as shown in our experiments.
%correlates well with the number of lexical simplifications \cite{xu2016optimizing}, but, as shown in Section \ref{qe-experiments}, this does not transfer as well to sentence-level %simplification 
%judgments.

\newcite{manchego2019easse} recently created a toolkit to calculate various standard automatic simplification metrics, %for assessing the quality of simplification system outputs, 
including SARI, word-level accuracy scores, and QE features such as compression ratio and average number of added/deleted words.
%There have been several recent works that have looked at
Recent work that addresses QE for simplification %These have experimented
experiment with a variety of features, including sentence length, average token length, and language model probabilities \cite{stajner2016quality,martin2018reference}. However, the best models from these works also use reference-reliant features such as BLEU and translation error rate, as these have been shown to correlate with Fluency and Adequacy. % \cite{martin2018reference}. 
%These works were done before the rise of large-scale pre-trained contextual embeddings, so these do not consider fine-tuning BERT, which Simple-QE explicitly leverages.
Note that these works were carried out before the rise of large-scale pre-trained models \cite{peters2018deep,devlin2019bert}. Sum-QE and our adaptation, Simple-QE, explicitly leverage the fine-tuning capabilities of BERT for assessing the quality of generated text.

\section{Methodology}
\label{methodology}
%\subsection{++}

To estimate the %overall 
quality of a simplification system output, we focus on three linguistic aspects:
% \begin{itemize}
% [noitemsep]
%     \item 
(a) \textbf{Fluency}: How well-formed %/grammatical 
    it is.
    %the system output 
    %\item 
    (b) \textbf{Adequacy}: How well  %the system output 
    it preserves the meaning of the original text. % sentence. \item 
    (c)   \textbf{Complexity}: How much 
    simpler it is than the original text. %sentence.
% \end{itemize} 
%In this work, 
We adapt the architecture proposed by \newcite{xenouleas2019sumqe} in their Sum-QE model, which extends the BERT fine-tuning process \cite{devlin2019bert}
%When creating our model, we make use of BERT, a large-scale model trained to produce high-quality contextual word embeddings \cite{devlin2019bert}. 
%Fine-tuning BERT has produced state-of-the-art results on many text classification tasks, including entailment prediction, sentiment analysis, and textual similarity, and gave very good results in the case of summarization.  %\cite{devlin2019bert}. 
%Building off of this generic fine-tuning framework, 
%Sum-QE 
to rate summaries with respect to five linguistic qualities. We expect Fluency, in our setting, %and Adequacy in the simplification case, 
to align well with Grammaticality as addressed by Sum-QE. % judgments.
%The three linguistic aspects used for measuring simplification quality can be seen as analogous to the five summarization qualities estimated by the SumQE model \cite{xenouleas2019sumqe}. Thus, one could argue that a straightforward application of this model could work. 
In the case of  Adequacy and Complexity, since judgments are relative (e.g. is the generated text {\it simpler than} the original text? does it %the simplified text 
convey {\it the same} meaning?), 
%of the original?),
we need to also consider the original complex text. % from which the simplification was derived. 

\newcite{xenouleas2019sumqe} use BERT as the main encoder and fine-tune it in three ways, one single-task %approach 
and two multi-task approaches:  %which leads to three versions of SUM-QE. %different approaches for predicting multiple qualities at the same time \cite{xenouleas2019sumqe}.

\begin{itemize} [noitemsep]
    \item \textbf{Single Task (S-1)}: Train $k$ models on each annotation type, where $k$ is the number of linguistic qualities; here, $k$ = 3.
    \item \textbf{Multi Task-1 (M-1)}: Train one model with a single regressor to predict $k$ annotations.
    \item \textbf{Multi Task-$k$ (M-3)}: Train one model with $k$ separate regressors. %corresponding to each individual annotation.
\end{itemize}

To adapt Sum-QE to %in estimating the quality of a generated 
simplification, we extend the architecture to take into account the original complex sentence. We do so by passing the original complex sentence and simplification system output through the BERT architecture separately. We concatenate the resulting embedding representations, and pass them through a final dense linear regressor layer $R$ to predict each linguistic quality score.
%the complexity score. Our extended architecture of the 
Our adaptation of the Sum-QE Multi Task-3 (M-3) approach 
%which integrates both the complex sentence and the system output, 
is described on the left side of Figure \ref{architecture}.

%Beyond this initial modification, 
To %better 
further adapt the QE model to our task, we also attempt to incorporate task-specific features: % These include: 
the average {\bf Length} of content words in a sentence in characters and in {\bf Syllables}, their {\bf Unigram Frequency},\footnote{We use the average log unigram frequency from the Google $n$-gram corpus \cite{thorsten2006web}.} the {\bf Sentence Length}, and the syntactic {\bf Parse Height}.
%We do so by passing 
We pass these features extracted from the original complex sentence separately through a linear layer, before concatenating them with the BERT embeddings of the sentence. We do the same for the system output. 
%system output.
%The architecture detailing this can be seen on 
The right side of Figure \ref{architecture} describes this architecture. 

\begin{comment}
\begin{itemize}
    \setlength\itemsep{1mm}
    \item \textbf{Word Length}: The average length of the content words (i.e. nouns, verbs, adjectives, and adverbs) in a sentence.
    \item \textbf{Syllables}: The average number of syllables of content words.
    \item \textbf{Unigram Frequency}: The average log unigram frequency of content words; the frequencies used here are from the Google $n$-gram corpus \cite{thorsten2006web}.
    \item \textbf{Sentence Length}: The average length of a sentence; this has been shown to be a strong baseline for predicting Complexity \cite{kriz2019complexity}.
    \item \textbf{Parse Height}: The height of the corresponding constituency parse tree for a sentence.
\end{itemize}
\end{comment}

\section{QE Experiments on System Output} \label{qe-experiments}

\subsection{Data and Baselines}
%\subsection{Datasets}

%For our test data, we utilize 
Our test data consists of human judgments collected by \newcite{kriz2019complexity} on %200 sentences from the Newsela corpus 
generated simplifications for 200 Newsela sentences \cite{xu2015problems}. For each sentence, outputs %outputs 
from six simplification models %systems %were 
were considered: vanilla Sequence-to-Sequence (Seq2Seq) \cite{nisioi2017exploring}, Seq2Seq with reinforcement learning \cite{zhang2017sentence}, memory-augmented Transformer \cite{zhao2018integrating}, and three variations of Seq2Seq with post-training re-ranking \newcite{kriz2019complexity}. Annotators were asked to rate the Fluency, Adequacy, and Complexity of each system output on a 5-point Likert Scale.\footnote{We do not use the QE dataset introduced by \newcite{stajner2016quality}, as it focuses on small-scale lexical changes -- similar to the Turk dataset \cite{xu2016optimizing} -- while current neural models %focus on 
adopt a more holistic approach.}
%This is because, similar to the Turk dataset \cite{xu2016optimizing}, in QATS %dataset %tends to put the focus is on small-scale lexical or phrasal changes in the output, while current neural models attempt to simplify a sentence more holistically.

%\subsection{Baselines} 

%In our initial experiments, 
We compare the Simple-QE model to %a variety of 
baselines that use %We again consider 
the simplification-specific features described in Section \ref{methodology}, quality estimates provided by BLEU \cite{papineni2002bleu} and SARI \cite{xu2016optimizing}, and three additional BERT-based baselines.  %We also include the several other baselines, each of which attempt to capture one or more aspect of simplification.

\begin{itemize}
    \item \textbf{BERT as Language Model (BERT LM)}: Given a sentence, we mask each token and predict the likelihood of the true word occurring in this context; this captures Fluency.\footnote{https://github.com/xu-song/bert-as-language-model}
    \item \textbf{BERT embedding similarity (BERT Sim)}: We convert the original and simplified texts %and the automatic simplification %reference and system output 
    into sentence-level BERT vector representations via mean pooling, and compute their cosine similarity; this estimates Adequacy. %This %attempts 
    %is expected to capture adequacy.
    \item \textbf{Sum-QE}: %Contrary to Simple-QE -- which has access to both the complex sentence and the system outputs -- our Sum-QE
    We apply Sum-QE directly, fine-tuning
    only on annotated system output. %s and %their 
    %the corresponding human judgment labels.
\end{itemize}

For Sum-QE and Simple-QE, we perform 10-fold cross validation, 
%fine-tuning on 90\% of the sentences, and testing on the held-out 10\%, 
combining the results to compute the overall correlation.

\subsection{Results}

The results are shown in Table \ref{SimpleQE-experiments}. %As we can see, 
Simple-QE correlates better with human judgments than the baseline models tested. The correlation of BLEU and SARI with human judgments is particularly low,  %somewhat low,
especially for Complexity.
%note that they have access to the reference simplification.
This is not surprising, given that SARI %has only been shown to correlate well with 
mainly addresses lexical simplification, %simplifications, 
while recent models approach simplification more holistically. 
%, which our work focuses on, try instead to generate a simplification in a more holistic manner.

The three versions of Simple-QE perform similar to Sum-QE on Fluency, where the model does not need to access the original complex sentence. The difference between the two models is more noticeable for Adequacy and Complexity, where accessing the original sentence actually helps Simple-QE make more reasonable %Adequacy and Complexity 
estimates. From the three versions of Simple-QE tested, the multi-task versions perform better than the single task on all three qualities tested.
The BERT LM and BERT Sim baselines perform well %regarding 
on Fluency and Adequacy, as expected, but fall short on the other aspects of simplification.

%We also %attempt to incorporate 
%test the impact of %the 
%simplification-specific features %mentioned in Section \ref{methodology} 
%on Simple-QE. The results %of this experiment 
%are shown in Table \ref{SimpleQE-features}. 
As shown in Table \ref{SimpleQE-experiments}, %the simplification-specific features 
adding numeric features do not improve performance.
%Surprisingly, these features do not seem to help %too
%much. This is likely the case 
This may be because the most predictive features, e.g. sentence length, are already implicitly learned by BERT.\footnote{The syntax of a sentence can be extracted from contextual word embeddings with reasonable accuracy \cite{hewitt2019structural}.}
%The feature that likely contains some new information is average word frequency; however, as we see in Section \ref{comp-pred}, this feature is not actually a good predictor of complexity level.
%predictive of complexity level.

\begin{table}
\begin{center}
\scalebox{0.90}{
\begin{tabular}{|c|c c c | } \hline
\textbf{Model} & \textbf{F} & \textbf{A} & \textbf{C} \\ \hline
BLEU & 0.183 & 0.305 & 0.057 \\
SARI & 0.133 & 0.207 & 0.013 \\ \hline
BERT LM & 0.411 & 0.350 & 0.249 \\
BERT Sim & 0.278 & 0.533 & 0.085 \\ \hline\hline
Sum-QE S-1 & 0.619 & 0.489 & 0.388 \\
Sum-QE M-1 & 0.632 & 0.541 & 0.407 \\
Sum-QE M-3 & 0.638 & 0.518 & 0.418 \\ \hline\hline
Simple-QE S-1 & 0.635 & 0.549 & 0.436 \\
Simple-QE M-1 & 0.643 & \textbf{0.630} & 0.433 \\
Simple-QE M-3 & \textbf{0.648} & 0.612 & \textbf{0.464} \\ \hline\hline
Simple-QE M-3 PH & \textbf{0.649} & 0.538 & 0.443 \\
Simple-QE M-3 SL & \textbf{0.650} & 0.541 & 0.432 \\
Simple-QE M-3 All & \textbf{0.650} & 0.538 & 0.436 \\ \hline
\end{tabular}
}
\end{center}
\caption{\label{SimpleQE-experiments} Pearson correlation with human  Fluency (F), Adequacy (A) and Complexity (C) judgments on simplification system output. The last three rows incorporate three numeric feature sets: Sentence Length (SL), Parse tree height (PH), and all features from Section \ref{methodology}.}
\end{table}

\section{Complexity Prediction on Human-written Text} \label{comp-pred}

\subsection{Datasets}

%Given that Simple-QE correlates somewhat worse with complexity, let's instead consider a simpler problem: 
As seen in the previous section, Simple-QE makes reasonable estimates for Fluency and Adequacy, but scores lower on Complexity. To explore this further, %we carry out an experiment where Simple-QE is asked to predict 
%we try to predict %a human written text's 
we test Sum-QE's capability to predict the Complexity of hand-written text. %This assumes 
Assuming  this text is relatively well-formed, we can in this way focus on how the model deals with Complexity. %isolate BERT's capability to predict on complexity, without worrying about %having to model % also  having to simultaneously predict factors such as grammaticality. and adequacy.

%This time, we use
We perform this analysis on the entire Newsela Corpus \cite{xu2015problems}, %Newsela
which contains 1,840 English news articles re-written at four complexity levels.\footnote{\url{https://newsela.com}} %We choose to train our model on this corpus because of its high quality and consistency. 
%In order to explore how well our complexity prediction model transfers to another domain, %outside of Newsela, 
%we perform an additional evaluation %out-of-domain evaluation %domain transfer experiments 
%on Parallel Wikipedia %texts 
%\cite{zhu2010monolingual}.%\footnote{The results of this experiment are given in the Appendix.} 
%\subsection{Experiments} \label{comp_experiments}
%In these experiments, 
We %compare 
explore how well fine-tuning BERT performs, compared to incorporating features into a linear regression classifier (LinReg).
Since we only address %are predictin  
Complexity, %the multi-task extensions of the models do not apply. 
we only consider the Single Task approach (Simple-QE S-1). %used by Sum-QE and Simple-QE do not apply. %are not necessary; thus, we only consider the Single Task approach (Sum-QE S-1), 

\subsection{Sentence-level Complexity Prediction} \label{sent-comp}

%In a first set of experiments, we attempt to predict sentence-level complexity. 
%To do this, 
We generate data by labelling sentences from Newsela articles with the complexity level of the corresponding document (0 to 4). 
%extract sentences from all Newsela articles, labeling each with the complexity level of its corresponding document. 
%Note that %is is possible for 
When a sentence is found at different complexity levels, we label it with the level of the simplest article in which it appears; this results in 370,376 sentences. 
%From there, for our feature baselines, we simply compute the correlation between the feature and each sentence's complexity label. 
We measure the correlation between each feature %presented in Section \ref{methodology} 
and the sentence's complexity level, and perform 10-fold cross-validation for LinReg and Simple-QE S-1. The results of this experiment are shown in the first column of Table \ref{comp-table}. %As we can see, 
Sentence length is the most predictive feature, and combining all the features using a linear regression classifier improves correlation, but both are outperformed by our model.
%This is somewhat unsurprising, since BERT has been shown to perform extremely well on a variety of text classification problems.

\subsection{Document-level Complexity Prediction} \label{doc-pred}

Most recent work has focused on sentence simplification, but document-level simplification might be more useful in practical settings. 
%if we want to build models that are more practically useful, we need to start considering document-level complexity. 
%To make document-level complexity predictions using Sum-QE S-1, 
%we consider two experimental setups. First, 
%taking the same sentence-level model training from the previous section, we can then 
%we average the complexity predictions made by our sentence-level model (Sum-QE S-1 Sent). 
%predict the complexity level of each sentence %individually, 
%using our sentence-level model  (Sum-QE S-1 Sent), and %then %simply
%take the average of these predictions. 
%One downside of this approach is that %these 
%sentences are %generally 
%relatively short %, with somewhat 
%and might have noisy labels, since not every sentence in an article is %at 
%assigned the same complexity level.
%We can 
%To alleviate these issues %if we instead attempt to 
%we experiment with making a single document-level complexity prediction. However, 
Given that BERT can only process 512 sub-word units at a time, %-- and most documents are longer than this -- %We thus 
we break a document down into sub-documents of up to 512 sub-word units. %; we do this by adding one paragraph at a time to a sub-document. 
At test time, to get a single document-level complexity prediction, we take a length-based weighted average of the predictions for the sub-documents. The results of this experiment are shown in the second column of Table \ref{comp-table}. We can see that, while sentence length and our linear regression classifier incorporating all features perform quite well, our model improves correlation even further.
%results in an even better document-level complexity prediction.

%We can see that combining BERT sub-document complexity predictions results in %an extremely a highly accurate document-level prediction, %again outperforming all %several simpler feature-based baselines. %In addition, Training on sub-documents improves this result even further. 
%This makes sense  because at the sentence level, there can sometimes be not enough complex signals of the sentence for the model to pick up on.
%partially due to the label ambiguity mentioned previously.
%However, at the document level there is generally more than enough complex (or simple) signals.

\begin{table}
\begin{center}
\scalebox{0.90}{
\begin{tabular}{|c|c|c|} \hline
\textbf{Model} & \textbf{Sentence} & \textbf{Document} \\ \hline
Word Length & 0.226 & 0.614 \\
Syllables & 0.235 & 0.614 \\
Frequency & 0.051 & 0.415 \\
Sentence Length & 0.548 & 0.907 \\
Parse Height & 0.341 & 0.795 \\
LinReg & 0.574 & 0.919 \\ \hline
Simple-QE S-1 & \textbf{0.726} & \textbf{0.964} \\ \hline
\end{tabular}
}
\end{center}
\caption{\label{comp-table} Pearson correlation of the  fine-tuned BERT model and different feature-based baselines on the Complexity prediction task at the sentence and the document-level.}
% fine-tuned BERT model, compared to several feature-based baselines and a linear regression combining the features (LinReg).}
\end{table}

\subsection{Out-of-Domain Evaluation}

%Our model's performance should also be able to transfer beyond Newsela. Thus, we also perform analyses where, after training on all Newsela sub-documents, we evaluate on aligned documents from PWKP \cite{zhu2010monolingual}.
Finally, we use PWKP \cite{zhu2010monolingual} to explore how well our complexity prediction model transfers to a different domain.
%adaptation of Sum-QE S-1 to predict human-written complexity our complexity prediction model transfers to a different domain. 
%Following \newcite{zhu2010monolingual}, 
We extract parallel texts from Simple and standard English Wikipedia dumps,\footnote{The Wikipedia dumps used can be found at https://dumps.wikimedia.org/simplewiki/ and https://dumps.wikimedia.org/enwiki/} aligning texts by their unique article ID. We label each Simple Wikipedia document with 0 (simpler) and each standard Wikipedia document with 1 (more complex). The alignment procedure results in 57,204 parallel documents. Similar to our document-level complexity prediction experiments, we combine the sentence and sub-document predictions into a single document-level prediction. In this experiment, our model reaches a Pearson correlation of 0.729. While this is lower than our in-domain Newsela experiments, it is still reasonably high, showing that our model can be applied for complexity prediction on other domains. 
%This experiment results in a Pearson correlation of 0.729, which is still reasonably high, although clearly lower results than on Newsela. This can likely be attributed to our labels being somewhat noisy, since not all Wikipedia articles are written at the same complexity level.

\section{Conclusion}

%In this paper, we present
We present Simple-QE, a quality estimation model for simplification. We have shown that extending Sum-QE \cite{xenouleas2019sumqe} to include the reference complex sentence sensibly improves %the quality of 
predictions on Adequacy and Complexity.
%significantly improves the correlation with adequacy and complexity human judgments. 
QE systems %have the potential to 
can be useful for evaluating the overall quality of model output without requiring expensive human annotations or references. Future simplification systems can incorporate Simple-QE into the optimization process, similar to how SARI was incorporated into a Seq2Seq network \cite{zhang2017sentence}. %We have also 
As shown, a fine-tuned BERT can also %is able to 
predict the complexity of human-written text, especially at the document level. This finding can be leveraged in future work as the simplification field moves towards simplifying entire documents. %multiple sentences or even entire documents at once.

\bibliography{naaclhlt2019}
\bibliographystyle{acl_natbib}

\end{document}

\maketitle

\section{Simple-QE}

In the main paper, we report Pearson correlations for our Simple-Quality Estimation (Simple-QE) and various simplification evaluation baselines. Here, we also show Spearman and Kendall correlations, in Table \ref{SimpleQE-experiments} and Table \ref{SimpleQE-features}, following \newcite{xenouleas2019sumqe}. Note that the three correlation metrics generally show the same patterns.

\section{Complexity Prediction}

Similarly, for our complexity prediction experiments, we also report Spearman and Kendall correlations in Tables \ref{sent-comp-table} and \ref{doc-comp-table}, though again they generally show the same trends. In addition, we show all three correlations for our out-of-domain experiments in Table \ref{wiki-comp-table}.

\begin{table}
\begin{center}
\begin{tabular}{|c|c|c|c|} \hline
\multirow{2}{*}{\textbf{Model}} & \multicolumn{3}{|c|}{\textbf{Correlation}} \\
& $\rho$ & $\tau$ & $r$ \\ \hline 
Word Length & 0.255 & 0.190 & 0.226 \\
Syllables & 0.263 & 0.198 & 0.235\\
Frequency & 0.092 & 0.067 & 0.051 \\
Sentence Length & 0.538 & 0.421 & 0.548 \\
Parse Height & 0.334 & 0.258 & 0.341 \\ 
LinReg & 0.573 & 0.442 & 0.574 \\ \hline
Sum-QE S-1 & \textbf{0.724} & \textbf{0.572} & \textbf{0.726} \\ \hline
\end{tabular}
\end{center}
\caption{\label{sent-comp-table} Correlations for the sentence-level complexity prediction fine-tuned BERT model, compared to several feature-based baselines, along with a linear regression combining the features (LinReg).}
\end{table}

\begin{table}
\begin{center}
\begin{tabular}{|c|c|c|c|} \hline
\multirow{2}{*}{\textbf{Model}} & \multicolumn{3}{|c|}{\textbf{Correlation}} \\
& $\rho$ & $\tau$ & $r$ \\ \hline 
Word Length & 0.619 & 0.472 & 0.614 \\
Syllables & 0.622 & 0.472 & 0.614 \\
Frequency & 0.425 & 0.317 & 0.415 \\
Sentence Length & 0.944 & 0.833 & 0.907 \\
Parse Height & 0.821 & 0.672 & 0.795 \\
LinReg & 0.948 & 0.838 & 0.919 \\ \hline
Sum-QE S-1 Doc & \textbf{0.961} & \textbf{0.860} & \textbf{0.964} \\ \hline
\end{tabular}
\end{center}
\caption{\label{doc-comp-table} Correlations for  document-level complexity prediction on the Newsela corpus.}
\end{table}

\begin{table}
\begin{center}
\begin{tabular}{|c|c|c|c|} \hline
\multirow{2}{*}{\textbf{Model}} & \multicolumn{3}{|c|}{\textbf{Correlation}} \\
& $\rho$ & $\tau$ & $r$ \\ \hline 
Sum-QE S-1 & \textbf{0.708} & \textbf{0.578} & \textbf{0.729} \\ \hline
\end{tabular}
\end{center}
\caption{\label{wiki-comp-table} Correlations for document-level complexity prediction when transferring our model trained on Newsela data to PWKP.}
%Correlations for complexity prediction at the document level, compared to several feature-based baselines.}
\end{table}

\begin{table*}
\begin{center}
\scalebox{0.90}{
\begin{tabular}{|c|c c c | c c c | c c c | } \hline
\multirow{2}{*}{\textbf{Model}} & \multicolumn{3}{|c|}{\textbf{Fluency}} & \multicolumn{3}{|c|}{\textbf{Adequacy}} & %\multicolumn{3}{|c|}{\textbf{Simplicity}}\\
\multicolumn{3}{|c|}{\textbf{Complexity}}\\
& $\rho$ & $\tau$ & $r$ & $\rho$ & $\tau$ & $r$ & $\rho$ & $\tau$ & $r$ \\ \hline
BLEU & 0.167 & 0.116 & 0.183 & 0.278 & 0.195 & 0.305 & 0.046 & 0.033 & 0.057 \\
SARI & 0.140 & 0.098 & 0.133 & 0.192 & 0.133 & 0.207 & 0.002 & 0.002 & 0.013 \\ \hline
BERT LM & 0.443 & 0.314 & 0.411 & 0.397 & 0.277 & 0.350 & 0.277 & 0.195 & 0.249 \\
BERT Sim & 0.288 & 0.202 & 0.278 & 0.530 & 0.382 & 0.533 & 0.095 & 0.066 & 0.085 \\ \hline
Sum-QE S-1 & 0.603 & 0.430 & 0.619 & 0.484 & 0.346 & 0.489 & 0.392 & 0.275 & 0.388 \\
Sum-QE M-1 & 0.626 & 0.446 & 0.632 & 0.535 & 0.384 & 0.541 & 0.403 & 0.282 & 0.407 \\
Sum-QE M-3 & 0.627 & 0.448 & 0.638 & 0.519 & 0.371 & 0.518 & 0.421 & 0.297 & 0.418 \\ \hline
Simple-QE S-1 & 0.619 & 0.443 & 0.635 & 0.545 & 0.393 & 0.549 & 0.438 & 0.309 & 0.436 \\
Simple-QE M-1 & 0.628 & 0.448 & 0.643 & \textbf{0.623} & \textbf{0.454} & \textbf{0.630} & 0.435 & 0.305 & 0.433 \\
Simple-QE M-3 & \textbf{0.638} & \textbf{0.459} & \textbf{0.648} & 0.604 & 0.439 & 0.612 & \textbf{0.459} & \textbf{0.325} & \textbf{0.464} \\ \hline
\end{tabular}
}
\end{center}
\caption{\label{SimpleQE-experiments} Correlations with human  Fluency, Adequacy and Complexity judgments on simplification system output; Spearman’s $\rho$, Kendall’s $\tau$ and Pearson’s $r$ are reported.}
\end{table*}

\begin{table*}
\begin{center}
\scalebox{0.90}{
\begin{tabular}{|c|c c c | c c c | c c c | } \hline
\multirow{2}{*}{\textbf{Model}} & \multicolumn{3}{|c|}{\textbf{Fluency}} & \multicolumn{3}{|c|}{\textbf{Adequacy}} & \multicolumn{3}{|c|}{\textbf{Complexity}}\\
& $\rho$ & $\tau$ & $r$ & $\rho$ & $\tau$ & $r$ & $\rho$ & $\tau$ & $r$ \\ \hline
Simple-QE M-3 PH & 0.641 & 0.460 & 0.649 & 0.535 & 0.384 & 0.538 & 0.451 & 0.317 & 0.443 \\
Simple-QE M-3 SL & 0.641 & 0.460 & 0.650 & 0.538 & 0.386 & 0.541 & 0.439 & 0.308 & 0.432 \\
Simple-QE M-3 All & 0.640 & 0.459 & 0.650 & 0.535 & 0.384 & 0.538 & 0.444 & 0.311 & 0.436 \\ \hline
\end{tabular}
}
\end{center}
\caption{\label{SimpleQE-features} Incorporating simplification-specific features into Simple-QE M-3, as this version performed the best on two of the three linguistic qualities. Feature sets include Sentence Length (SL), Parse Height (PH), and all features discussed in the main paper.}
\end{table*}

\bibliography{naaclhlt2019}
\bibliographystyle{acl_natbib}